\documentclass{article}
\usepackage{ICASSP2021,amsmath,graphicx}
\usepackage{booktabs}
\usepackage{multirow} 
\usepackage{hhline} 
\usepackage{pifont}
\usepackage{hyperref} 
\newcommand{\cmark}{\ding{51}}%
\newcommand{\xmark}{\ding{55}}%

\usepackage{xcolor}

\usepackage[normalem]{ulem}

\usepackage[backend=bibtex,citestyle=numeric-comp,bibstyle=ieee,sorting=none,defernumbers=true,giveninits=true,doi=false,isbn=false,url=false,eprint=false,minbibnames=8,maxbibnames=8]{biblatex}
\usepackage{subcaption}

\title{Towards practical lipreading with distilled and efficient models}
\name{Pingchuan Ma$^{\dagger,1}$, Brais Martinez$^{\dagger,2}$, Stavros Petridis$^{1,2}$, Maja Pantic$^1$}
\address{
  $^1$Department of Computing, Imperial College London, UK\\
  $^2$Samsung AI Research Center, Cambridge, UK}

\begin{document}

\maketitle

\begin{abstract}
Lipreading has witnessed a lot of progress due to the resurgence of neural networks. Recent works have placed emphasis on aspects such as improving performance by finding the optimal architecture or improving generalization. However, there is still a significant gap between the current methodologies and the requirements for an effective deployment of lipreading in practical scenarios. In this work, we propose a series of innovations that significantly bridge that gap: first, we raise the state-of-the-art performance by a wide margin on LRW and LRW-1000 to 88.5\,\% and 46.6\,\%, respectively using self-distillation. Secondly, we propose a series of architectural changes, including a novel Depthwise  Separable Temporal Convolutional Network (DS-TCN) head, that slashes the computational cost to a fraction of the (already quite efficient) original model. Thirdly, we show that knowledge distillation is a very effective tool for recovering performance of the lightweight models. This results in a range of models with different accuracy-efficiency trade-offs. However, our most promising lightweight models are on par with the current state-of-the-art while showing a reduction of 8.2$\times$ and 3.9$\times$ in terms of computational cost and number of parameters, respectively, which we hope will enable the deployment of lipreading models in practical applications.
\end{abstract}

\let\thefootnote\relax\footnote{$\dagger$ The first two authors contributed equally.}

\section{Introduction}
\label{sec:intro}

Lipreading has attracted a lot of research interest recently thanks to the superior performance of deep architectures. Such models consist of either fully connected \cite{petridis2017deepVisualSpeech,petridis2018visualWhisper, petridis2017end,wand16,petridis17AV} or convolutional layers~\cite{stafylakis17, shillingford2018large, afouras2018deep, chung16b} which extract features from the mouth region of interest, followed by recurrent layers or attention \cite{chung16b,petridis2018audio} / self-attention architectures \cite{afouras2018deep}. However, one of the major limitations of current models barring their use in practical applications is  their computational cost. Many speech recognition applications rely on on-device computing, where the computational capacity is limited, and memory footprint and battery consumption are also important factors. As a consequence, few works have also focused on the computational complexity of visual speech recognition \cite{koumparoulis2019mobilipnet, shrivastava2019mobivsr}, but such models still trail massively behind full-fledged ones in terms of accuracy.

\looseness - 1

In this work we focus on improving the performance of the state-of-the-art model and training lightweight models without considerable decrease in performance.  Lipreading is a challenging task due to the nature of the signal, where a model is tasked with distinguishing between e.g. \textit{million} and \textit{millions} solely based on visual information. We resort to Knowledge Distillation (KD)~\cite{hinton2014distilling} since it provides an extra supervisory signal with inter-class similarity information. For example, if two classes are very similar as in the case above, the KD loss will penalize less when the algorithm confuses them. We leverage this insight to produce a sequence of teacher-student classifiers in the same manner as \cite{furlanello2018born, yang2018knowledge}, by which student and teacher have the same architecture, and the student will become the teacher in the next generation until no improvement is observed.

Our second contribution is to propose a novel lightweight architecture. The ResNet-18 backbone can be readily exchanged for an efficient one, such as a version of the MobileNet~\cite{howard2017mobilenets} or ShuffleNet~\cite{ma2018shufflenet} families. However, there is no such equivalent for the head classifier. The key to designing the efficient backbones is the use of depthwise separable convolutions (a depthwise convolution followed by a pointwise convolution) \cite{chollet2017xception} to replace standard convolutions. This operation dramatically reduces the amount of parameters and the number of FLOPs. Thus, we devise a novel variant of the Temporal Convolutional Networks that relies on depthwise separable convolutions instead. 

Our third contribution is to use the KD framework to recover some of the performance drop of these efficient networks. Unlike the full-fledged case, it is now possible to use a higher-capacity network to drive the optimization. However, we find that just using the best-performing model as the teacher, which is the standard practice in the literature, yields sub-optimal performance. Instead, we use intermediate networks whose architecture is in-between the full-fledged and the efficient one. Thus, similar to \cite{Martinez2020Training}, we generate a sequence of teacher-student pairs that progressively bridges the architectural gap.

We provide experimental evidence showing that a) we achieve new state-of-the-art performance on LRW \cite{chung16b} and LRW-1000 \cite{lrw1000} by a wide margin and without any increase of computational performance$^1$ \footnote{$^1$ The models and code are available at \url{https://sites.google.com/view/audiovisual-speech-recognition}} and b) our lightweight models can achieve competitive performance. For example, we match the current state-of-the-art on LRW \cite{martinez2020lipreading} using 8.2$\times$ fewer FLOPs and 3.9$\times$ fewer parameters.

\begin{figure}[t]
    \centering
    \includegraphics[width=\columnwidth]{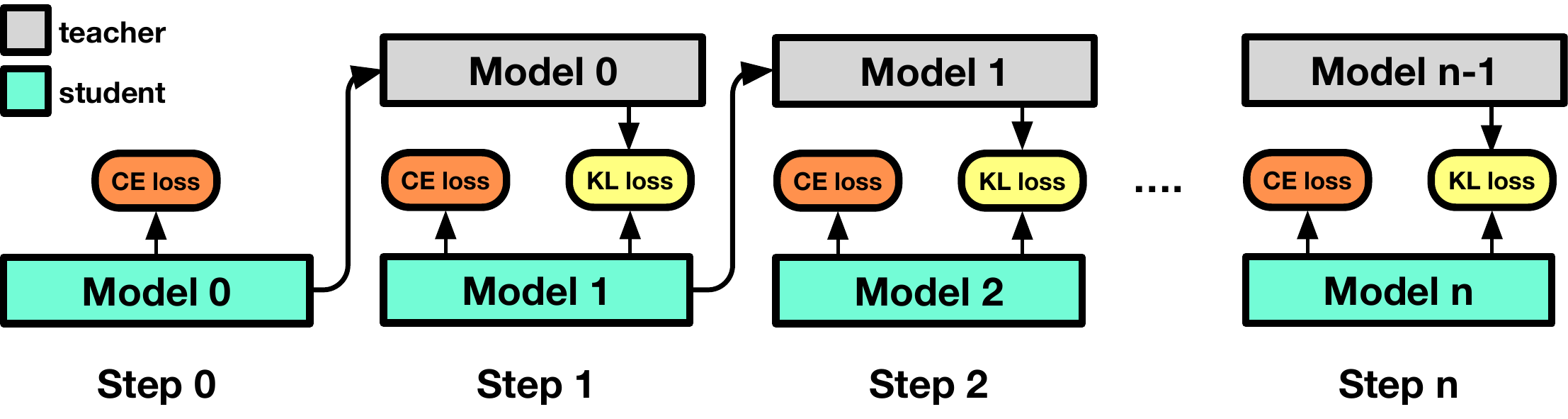}
    \caption[architecture]{The pipeline of knowledge distillation in generations}
\label{fig:pipeline}
\vspace{-2mm}
\end{figure}

\section{Towards Practical Lipreading}
\label{sec:Architecture}
\begin{figure}[t]
    \centering
    \begin{subfigure}{.22\columnwidth}
    \includegraphics[width=\linewidth]{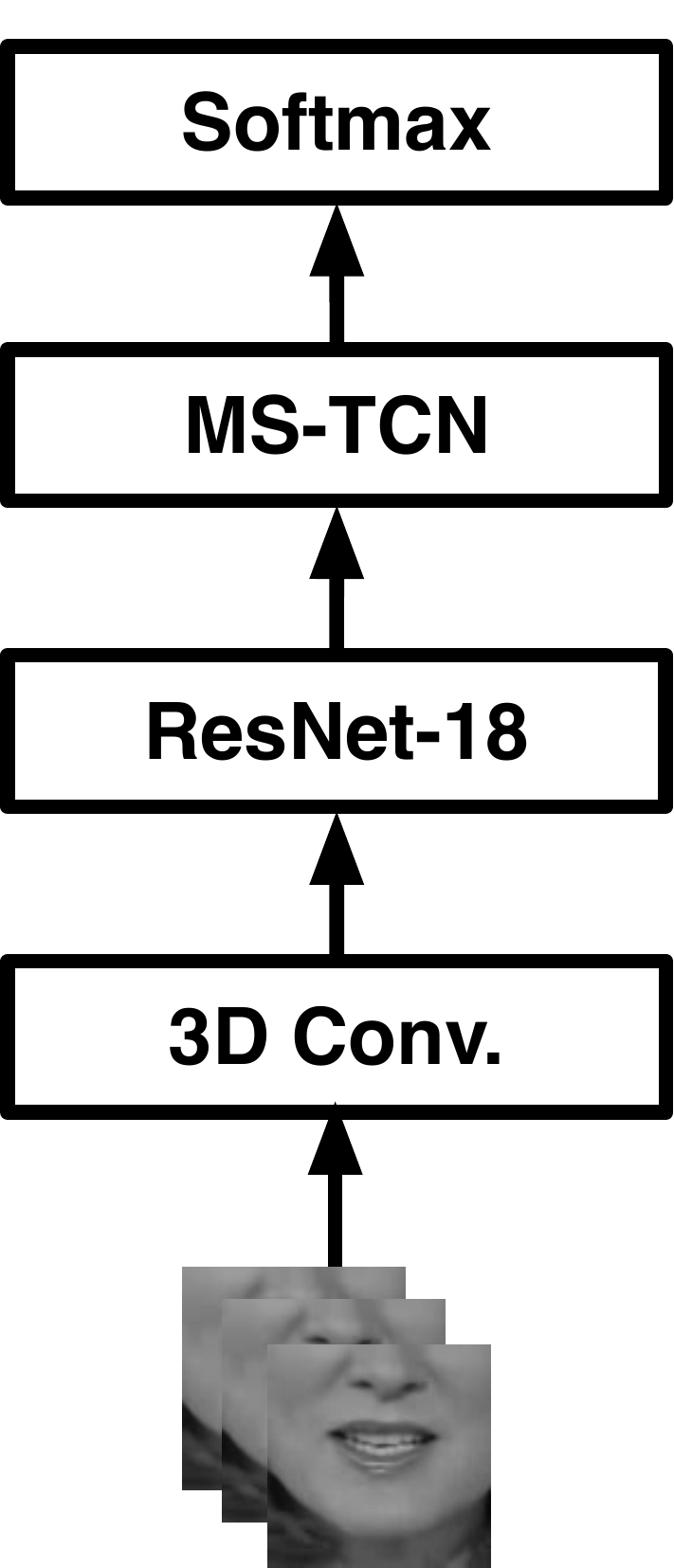}
    \caption{}
    \label{fig:variations_a}  
    \end{subfigure}\hfill%
    \begin{subfigure}{.22\columnwidth}
    \includegraphics[width=\linewidth]{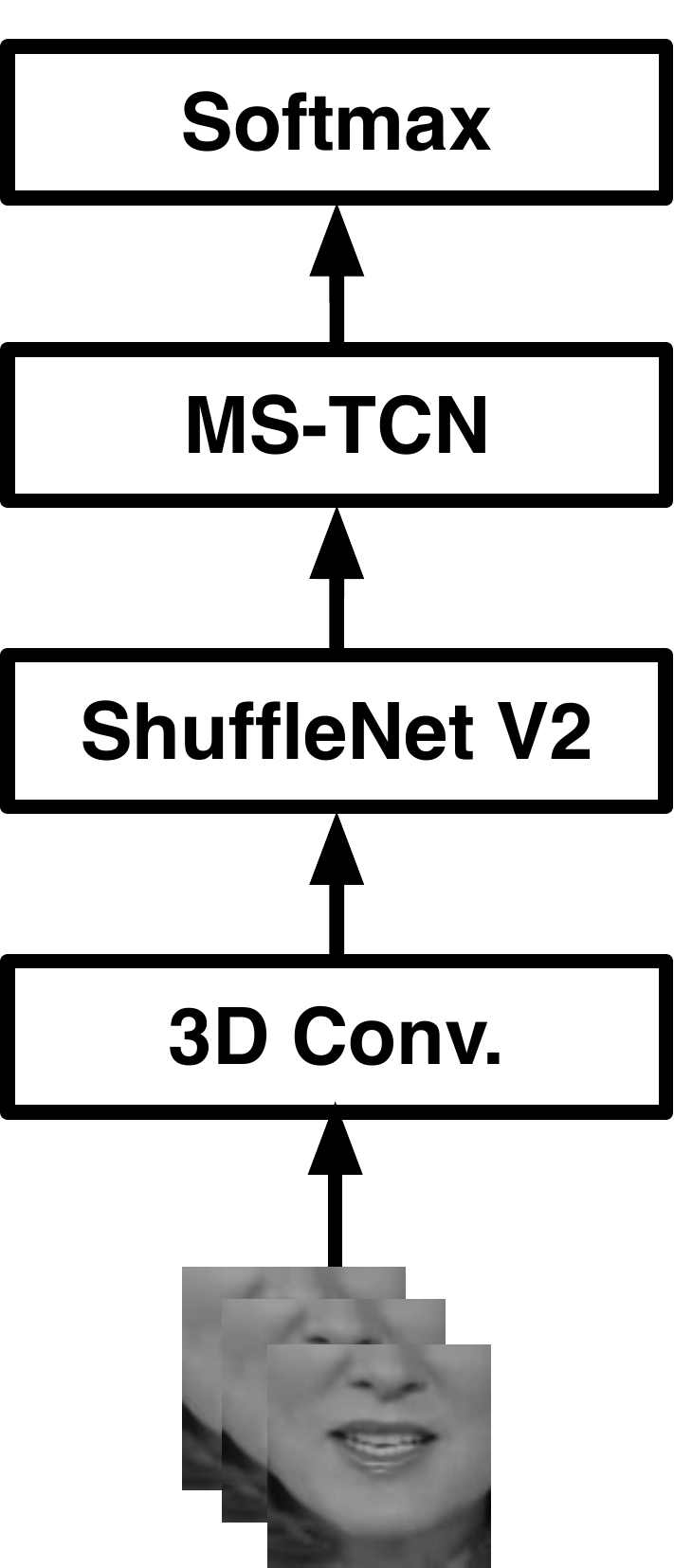}
    \caption{}
    \label{fig:variations_b}  
    \end{subfigure}\hfill%
    \begin{subfigure}{.22\columnwidth}
    \includegraphics[width=\linewidth]{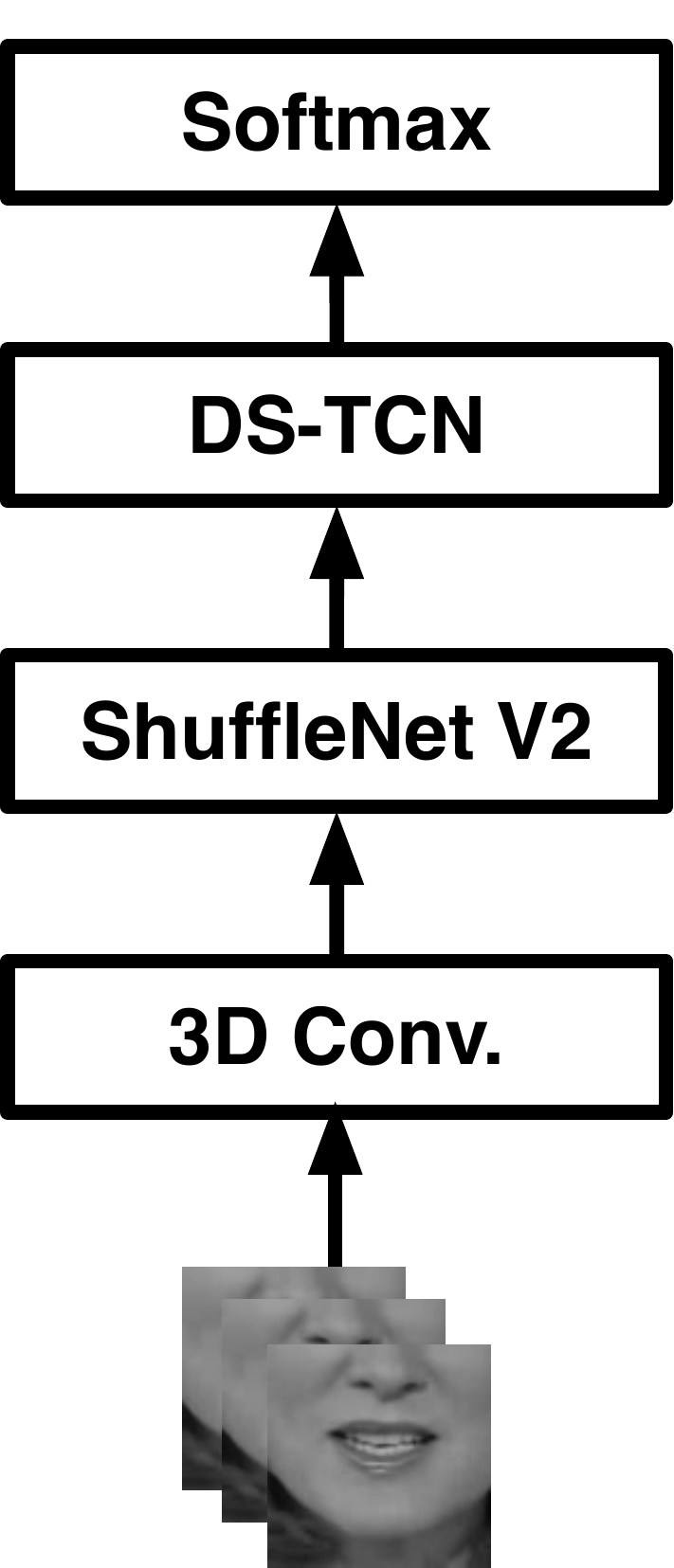}
    \caption{}
    \label{fig:variations_c}  
    \end{subfigure}
    
    \caption[architecture]{(a): Base architecture with ResNet\-18 and multi-scale TCN, (b): Lipreading model with ShuffleNet v2 backbone and multi-scale TCN back-end. (c): Lipreading model with ShuffleNet v2 backbone and depthwise separable TCN back-end.
    }
\label{fig:architecture}
\vspace{-2mm}
\end{figure}

\looseness - 1
\noindent\textbf{Base Architecture}\quad
We use the visual speech recognition architecture proposed in \cite{martinez2020lipreading} as our base architecture  which achieves the state-of-the-art performance on the LRW and LRW1000 datasets. The details are shown in Fig.~\ref{fig:variations_a}. It consists of a modified ResNet-18 backbone in which the first convolution has been substituted by a 3D convolution of kernel size $5\times 7\times 7$. The rest of the network follows a standard design up to the global average pooling layer. A multi-scale temporal convolutional network (MS-TCN) follows to model the short-term and long-term temporal information simultaneously.

\noindent\textbf{Efficient Backbone}\quad
The efficient spatial backbone is produced by replacing the ResNet-18 with an efficient network based on depthwise separable convolutions. For the purpose of this study, we use ShuffleNet v2 ($\beta\times$) as the backbone, where $\beta$ is the width multiplier \cite{ma2018shufflenet}. This architecture uses depthwise convolutions and channel shuffling which is designed to enable information communication between different groups of channels. ShuffleNet v2 (1.0$\times$) has 5$\times$ fewer parameters and 12$\times$ fewer FLOPs than ResNet-18. The architecture is shown in Fig.~\ref{fig:variations_b}.

\noindent\textbf{Depthwise Separable TCN}\quad
We note that the cost of the convolution operation with kernel size greater than 1 in MS-TCN is non-negligible. To build an efficient architecture (shown in Fig.~\ref{fig:variations_c}), we replace standard convolutions with depthwise separable convolutions in MS-TCN. We first apply in each channel a convolution with kernel size $k$, where channel interactions are directly ignored. This is followed by a point-wise convolution with kernel size 1 which transforms the $C_{in}$ input channels  to $C_{out}$ output channels.  Thus, the cost of convolution is reduced from $k \times C_{in} \times C_{out}$ (standard convolution) to $ k\times C_{in} + C_{in}C_{out}$. The architecture is denoted as a Depthwise Separable Temporal Convolutional Network (DS-TCN). 

\noindent\textbf{Knowledge Distillation}\quad
Knowledge Distillation (KD) \cite{hinton2014distilling} was initially proposed to transfer knowledge from a teacher model to a student model for compression purposes, i.e., the student capacity is much smaller than the teacher one. Recent studies \cite{furlanello2018born, bagherinezhad2018label, yang2018knowledge} have experimentally shown that the student can still benefit when the teacher and student network have identical architectures. This naturally gave rise to the idea of training in generations. In particular, the student of one generation is used as the teacher of the subsequent generation. This self-distillation process, called born-again distillation, is iterated until no further improvement is observed. Finally, an ensemble can be optionally used so as to combine the predictions from multiple generations \cite{furlanello2018born}. The training pipeline is shown in Fig.~\ref{fig:pipeline}.

In this work, we use born-again distillation for improving the performance of the state-of-the-art model. We also use the standard knowledge distillation to train a series of efficient models where each student has smaller capacity than the teacher. In both cases, we aim to minimise the combination of cross-entropy loss ($\mathcal{L}_{{\rm CE}}$) for hard targets and Kullback-Leibler (KL) divergence loss ($\mathcal{L}_{{\rm KD}}$) for soft targets. Let us denote the labels as $y$, the parameters of student and teacher models as $\theta_s$ and $\theta_t$, respectively, and the predictions from the student and teacher models as $z_s$ and $z_t$, respectively. $\delta(\cdot)$ denotes the softmax function and $\alpha$ is a hyperparameter to balance the loss terms. The overall loss function is calculated as follows:
\begin{equation}
\mathcal{L} = \mathcal{L}_{{\rm CE}}(y, \delta(z_s;\theta_s)) + \alpha \mathcal{L}_{{\rm KD}}(\delta(z_s;\theta_s), \delta(z_t;\theta_t))
\label{eq:KD_loss}
\end{equation}

Note that we have omitted the temperature term, which is commonly used to soften the logits of the $\mathcal{L}_{{\rm KD}}$ term, since we found it to be unnecessary in our case.

\section{Experimental Setup}

\noindent\textbf{Datasets}\quad
Lip Reading in the Wild (LRW) \cite{chung16b} is based on a collection of over 1000 speakers from BBC programs.
It contains over half a million utterances of 500 English words. Each utterance is composed of 29 frames (1.16 seconds), where the target word is surrounded by other context words. 
LRW-1000 \cite{lrw1000} contains more than 2000 speakers and has 1000 Mandarin syllable-based classes with a total of 718018 utterances. It contains utterances of varying length from 0.01 up to 2.25 seconds.
\begin{table}[t]
\begin{center}\scalebox{.85}{
\renewcommand{\arraystretch}{1.05}
\begin{tabular}{lcc}
\toprule
Method  & \multicolumn{2}{c}{Top-1 Acc. (\%)}  \\
\midrule
3D-CNN \cite{chung16}  & \multicolumn{2}{c}{61.1}  \\
Seq-to-Seq \cite{chung16b} &  \multicolumn{2}{c}{76.2} \\
ResNet34 + BLSTM \cite{stafylakis17}  & \multicolumn{2}{c}{83.0} \\
ResNet34 + BGRU \cite{petridis18} & \multicolumn{2}{c}{83.4} \\
2-stream 3D-CNN + BLSTM \cite{weng19}  & \multicolumn{2}{c}{84.1} \\
ResNet-18 + BLSTM \cite{stafylakis2018pushing} & \multicolumn{2}{c}{84.3}  \\
ResNet-18 + BGRU + Cutout \cite{zhang2020can} & \multicolumn{2}{c}{85.0}\\
ResNet-18 + MS-TCN \cite{martinez2020lipreading} & \multicolumn{2}{c}{85.3}  \\
\midrule
ResNet-18 + MS-TCN - Teacher  & \multicolumn{2}{c}{85.3} \\ 
ResNet-18 + MS-TCN - Student 1 & \multicolumn{2}{c}{87.4} \\ 
ResNet-18 + MS-TCN - Student 2 & \multicolumn{2}{c}{87.8}  \\ 
ResNet-18 + MS-TCN - Student 3 & \multicolumn{2}{c}{\textbf{87.9}} \\
ResNet-18 + MS-TCN - Student 4 & \multicolumn{2}{c}{87.7} \\ 
Ensemble & \multicolumn{2}{c}{\textbf{88.5}} \\
\bottomrule
\end{tabular}}
\vspace{-3mm}
\end{center}
\caption{Comparison with state-of-the-art methods on the LRW dataset in terms of classification accuracy. Each student is trained using the model from the line above as a teacher. Student $i$ stands for the model after the $i$-th self-distillation iteration.}
\label{tab:sota_visual_lrw}
\vspace{-4mm}
\end{table}

\noindent\textbf{Pre-processing}\quad
For the video sequences in the LRW dataset, 68 facial landmarks are detected and tracked using dlib \cite{king2009dlib}. The faces are aligned to a neural reference frame and a bounding box of $96\times96$ is used to crop the mouth region of interest. Video sequences in the LRW1000 dataset are already cropped so there is no need for pre-processing.

\looseness-1
\noindent\textbf{Training}\quad
We use the same training parameters as \cite{martinez2020lipreading}. The only exception is the use of Adam with decoupled Weight decay (AdamW) \cite{loshchilov2019decoupled} with $\beta_{1}=0.9$, $\beta_{2}=0.999$, $\epsilon=10^{-8}$ and a weight decay of 0.01. The network is trained for 80 epochs using an initial learning rate of 0.0003, and a mini-batch of 32. We decay the learning rate using a cosine annealing schedule without warm-up steps. All models are randomly initialised and no external datasets are used.

\looseness - 1
\noindent\textbf{Data Augmentation}\quad
During training, each sequence is flipped horizontally with a probability of 0.5, randomly cropped to a size of $88\times88$ and mixup \cite{zhang2017mixup} is used with a weight of 0.4. During testing, we use the $88\times88$ center patch of the image sequence. 
To improve robustness, we train all models with variable-length augmentation similarly to \cite{martinez2020lipreading}.

\section{Results}
\label{sec:results}

\begin{table}
\begin{center}\scalebox{.85}{
\renewcommand{\arraystretch}{1.05}
\begin{tabular}{lcc}
\toprule
Method  &\multicolumn{2}{c}{Top-1 Acc. (\%)}  \\ \hline
ResNet34 + DenseNet52 + ConvLSTM \cite{wang2019} & \multicolumn{2}{c}{36.9} \\
ResNet34 + BLSTM \cite{stafylakis17} &\multicolumn{2}{c}{38.2} \\
ResNet-18 + BGRU  \cite{zhang2020can}  &\multicolumn{2}{c}{38.6}\\ 
ResNet-18 + MS-TCN \cite{martinez2020lipreading}  &\multicolumn{2}{c}{41.4} \\
ResNet-18 + BGRU + Cutout \cite{zhang2020can} &\multicolumn{2}{c}{45.2 $\dagger$}\\ 
\midrule
ResNet-18 + MS-TCN - Teacher & \multicolumn{2}{c}{43.2} \\ 
ResNet-18 + MS-TCN - Student 1& \multicolumn{2}{c}{\textbf{45.3}} \\
ResNet-18 + MS-TCN - Student 2 & \multicolumn{2}{c}{44.7} \\
Ensemble & \multicolumn{2}{c}{\textbf{46.6}} \\ \bottomrule
\end{tabular}}
\vspace{-3mm}
\end{center}
\caption{Comparison with state-of-the-art methods on the LRW-1000 dataset in terms of classification accuracy using the publicly available version of the database (which provides the cropped mouth regions). Each student is trained using the model from the line above as a teacher. $\dagger$ \cite{zhang2020can} uses the full face version of the database, which is not publicly available. Student $i$ stands for the model after the $i$-th self-distillation iteration.}
\label{tab:sota_visual_lrw1000}
\vspace{-4mm}
\end{table}

\noindent\textbf{Born-Again Distillation}\quad
In here we show that adding a distillation loss adds valuable inter-class similarity information and in turn helps the optimization procedure. Thus, we resort to the self-distillation strategy (born-again distillation), e.g.~\cite{furlanello2018born,bagherinezhad2018label}, where both the teacher and the student networks have the same architecture, as explained in section \ref{sec:Architecture}.

\looseness - 1
Results on the LRW dataset are shown in Table. \ref{tab:sota_visual_lrw}. This strategy leads to a new state-of-the-art performance  by 2.6\,\% margin over the previous one without any computational cost increase. Furthermore, an ensemble of models, a strategy suggested in \cite{furlanello2018born}, reaches an accuracy of 88.5\,\%, which further pushes the state-of-the-art performance on LRW.

\looseness-1
Results on the LRW-1000 dataset are shown in Table \ref{tab:sota_visual_lrw1000}. In this case, our proposed best single-model and ensemble result in an absolute improvement of 3.9\,\% and 5.2\,\%,  respectively, over the state-of-the-art accuracy. We should note that we only compare with works which use the publicly available version of the database. These results confirm that adding inter-class similarity information is  useful for lipreading.

\begin{table}
\begin{center}{\scalebox{0.67}{
\renewcommand{\arraystretch}{1.1}
\begin{tabular}{llcccc}
\toprule
Student Backbone & Student Back-end & Distillation & Top-1 & Params & FLOPs\\
(Width mult.)&(Width mult.) & & Acc. & $\times10^6$ &$\times10^9$ \\
\midrule
ResNet-18 \cite{martinez2020lipreading} & MS-TCN (3$\times$) & - & 85.3 & 36.4 &10.31 \\
ResNet-34 \cite{petridis18}        & BGRU (512)              & - & 83.4 & 29.7 & 18.71 \\ 
MobiVSR-1 \cite{shrivastava2019mobivsr} & TCN                & - & 72.2 & 4.5 & 10.75 \\
\midrule
\multirow{2}*{ShuffleNet v2 (1$\times$)} & MS-TCN (3$\times$) & \xmark & 84.4 & 28.8 & 2.23 \\
                                         & MS-TCN (3$\times$) & \cmark & 85.5 &28.8 &2.23 \\
\midrule
\multirow{2}*{ShuffleNet v2 (1$\times$)} & DS-MS-TCN (3$\times$) & \xmark & 84.4 & 9.3 & 1.26 \\
                                         & DS-MS-TCN (3$\times$) & \cmark & 85.3 & 9.3 & 1.26 \\ 
\midrule
\multirow{2}*{ShuffleNet v2 (1$\times$)} & TCN (1$\times$) & \xmark & 81.0 & 3.8 & 1.12 \\
                                         & TCN (1$\times$) & \cmark & 82.7 & 3.8 &1.12 \\ 
\midrule
\multirow{2}*{ShuffleNet v2 (0.5$\times$)} & TCN (1$\times$) & \xmark & 78.1 & 2.9 & 0.58
\\
                                           & TCN (1$\times$) & \cmark & 79.9 & 2.9 & 0.58 \\

\bottomrule
\end{tabular}}}
\end{center}
\vspace{-3mm}
\caption{Performance of different efficient models, ordered in descending computational complexity, and their comparison to the state-of-the-art on the LRW dataset. We use a sequence of 29-frames with a size of 88 by 88 pixels to compute the multiply-add operations (FLOPs). The number of channels is scaled for different capacities, marked as $0.5 \times$, $1 \times$, and $2 \times$. Channel widths are the standard ones for ShuffleNet~V2, while base channel width for TCN is 256 channels.}
\label{tab:efficient_visual_lw_lrw500}
\vspace{-4mm}
\end{table}

\noindent\textbf{Efficient Lipreading}\quad
The frame encoder can be made more efficient by replacing the ResNet-18 with a lightweight ShuffleNet v2 (shown in Fig.~\ref{fig:variations_b}), as explained in section \ref{sec:Architecture}. We should note that we maintain the first convolution of the network as a 3D convolution.  Preliminary experiments showed ShuffleNet~v2~\cite{ma2018shufflenet} yields superior performance over other lightweight architectures like MobileNetV2~\cite{sandler2018mobilenetv2}. It can be seen in Table \ref{tab:efficient_visual_lw_lrw500} that this change results in a drop of 0.9\,\% while reducing both the number of parameters and FLOPs.

\looseness-1
The next step is the replacement of the MS-TCN head with its depthwise-separable variant, noted as DS-MS-TCN. As shown in Table \ref{tab:efficient_visual_lw_lrw500} this variant leads to a model with almost one third of parameters and a 50\,\% reduction in FLOPs while achieving the same  accuracy as the ShuffleNet v2 with a MS-TCN head.

\looseness - 1
Models can become even lighter (shown in Fig.~\ref{fig:variations_c}) by reducing the number of heads to 1, denoted by TCN,  and by reducing the width multiplied of the ShuffleNet v2 to 0.5. In the former case, performance drops by 3.4\,\%, and in the latter by a further 1.9\,\% resulting in accuracy of 78.1\,\%. However, it should be noted that the number of parameters and FLOPs is significantly reduced for both models.

\looseness - 1
Results for LRW-1000 can be seen in Table \ref{tab:efficient_visual_lw_lrw1000}. In this case the use of a ShuffleNet v2 with single TCN head leads to a small drop of 0.7\,\% compared to the full model while reducing significantly the number of parameters and FLOPs. In addition, the use of DS-TCN results in a further drop of 1.6\,\%. It is also interesting to note that the use of  ShuffleNet v2 with a width multiplier of 0.5 achieves the same performance as the baseline ShuffleNet v2.

\noindent\textbf{Sequential Distillation}\quad In order to partially bridge this gap, we explore Knowledge Distillation once again. Since now there are higher capacity models that can act as teachers, we do not need to resort to self distillation. We first explored the standard distillation approach in which we take the best-performing model as the teacher. However, it is known that a wider gap in terms of architecture might mean a less effective transfer~\cite{search_to_distill_cvpr20,Martinez2020Training}. Thus, we also explore a sequential distillation approach. More specifically, for lower-capacity networks, we use intermediate-capacity networks to more progressively bridging the architectural gap. For example, for the ShuffleNet~v2 (1$\times$)+DS-MS-TCN, we can first train a model using the full fledged ResNet-18+MS-TCN model as teacher, and use the ShuffleNet~v2 (1$\times$)+MS-TCN as the student. Then, on the second step, we use the latter model as the teacher, and train our target model, ShuffleNet~v2 (1$\times$)+DS-MS-TCN, as the student. This procedure resembles the self-distillation strategy described above in the sense that trains a sequence of teacher-student pairs, where the previous student becomes the teacher in the next iteration. However, unlike that strategy, it progressively changes the architecture from the full-fledged model to the target architecture.

\looseness - 1
The results on the LRW dataset are shown in Table \ref{tab:efficient_visual_lw_lrw500}. Replacing the state-of-the-art ResNet-18+MS-TCN with ShuffleNet~v2 (1$\times$)+ DS-MS-TCN leads to the same accuracy, after distillation, than the previous state-of-the-art MS-TCN of \cite{martinez2020lipreading}, while requiring 8.2$\times$ fewer FLOPs and 3.9$\times$ fewer parameters. This is a significant finding since the MS-TCN is already quite efficient, having slightly lower computational cost than the lightweight architecture of MobiVSR-1 \cite{shrivastava2019mobivsr}. Another interesting combination is the ShuffleNet~v2 (0.5$\times$)+ TCN model, which achieves 79.9\,\% accuracy on LRW with as little as 0.58G FLOPs and 2.9M parameters, a reduction of 17.8$\times$ and 12.5$\times$ respectively when compared to the ResNet-18+MS-TCN model of \cite{martinez2020lipreading}.

\looseness - 1
The same pattern is also observed on the LRW1000 dataset, which is shown in Table~\ref{tab:efficient_visual_lw_lrw1000}. ShuffleNet~v2 (0.5$\times$)+DS-TCN ($1\times$) after distillation results in a drop of 1.2\% in accuracy,  while requiring 22.9$\times$ fewer parameters and 18.8$\times$ fewer FLOPs than the state-of-the-art-model. \cite{martinez2020lipreading}.

\begin{table}
\begin{center}{\scalebox{0.68}{
\renewcommand{\arraystretch}{1.1}
\begin{tabular}{llcccc}
\toprule
Student Backbone & Student Back-end & Distillation & Top-1 &Params &FLOPs\\
(Width mult.) & (Width mult.) &  &Acc. &$\times10^6$ &$\times10^9$ \\\midrule
ResNet-18 \cite{martinez2020lipreading} & MS-TCN(3$\times$) & - & 41.4 &36.7 &15.78 \\
3D DenseNet \cite{lrw1000} &BGRU (256) &- &34.8 &15.0 & 30.32\\ 
\midrule
\multirow{2}*{ShuffleNet v2 (1$\times$)} &TCN (1$\times$) & \xmark &40.7 &3.9 &1.73  \\
                                         &TCN (1$\times$) & \cmark &41.4 &3.9 &1.73 \\
\midrule
\multirow{2}*{ShuffleNet v2 (1$\times$)} & DS-TCN (1$\times$) & \xmark &39.1 &2.5 &1.68 \\
                                         & DS-TCN (1$\times$) & \cmark &40.4 &2.5 &1.68 \\
\midrule
\multirow{2}*{ShuffleNet v2 (0.5$\times$)} & TCN (1$\times$) & \xmark &40.5 &3.0 &0.89 \\
                                           & TCN (1$\times$) & \cmark &41.1 &3.0 &0.89 \\
\midrule
\multirow{2}*{ShuffleNet v2 (0.5$\times$)} & DS-TCN (1$\times$) & \xmark & 39.1 &1.6 &0.84 \\
                                           & DS-TCN (1$\times$) & \cmark &40.2 &1.6 &0.84 \\
\bottomrule
\end{tabular}}}

\end{center}
\vspace{-3mm}
\caption{Performance of different efficient models on the LRW-1000 dataset. We use a sequence of 29-frame with a size of 112 by 112 to report multiply-add operations (FLOPs). The number of channels is scaled for different capacities, marked as $0.5 \times$ and $1 \times$. Channel widths are the standard ones for ShuffleNet v2, while base channel width for TCN is 256 channels.}
\label{tab:efficient_visual_lw_lrw1000}
\vspace{-4mm}
\end{table}

\section{Conclusions}
\looseness - 1
In this work, we present state-of-the-art results on isolated word recognition by knowledge distillation. We also investigate efficient models for visual speech recognition and we achieve results similar to the current state-of-the-art while reducing the computational cost by 8 times. It would be interesting to investigate in future work how cross-modal distillation affects the performance of audiovisual speech recognition models.

\clearpage
\AtNextBibliography{\small}
\section{References}
\begingroup
\setlength\bibitemsep{1pt}
\printbibliography[heading=none]
\endgroup
\end{document}